\documentclass[10pt,a4paper]{article}

\usepackage{geometry}
\newgeometry{ hmargin={25mm,25mm}}

\usepackage{algorithm}
\usepackage[noend]{algpseudocode}
\usepackage{amsmath}
\usepackage{amssymb}
\usepackage{bm}
\usepackage{booktabs}
\usepackage{caption,subcaption}

\usepackage{multicol}
\usepackage[numbers]{natbib}
\usepackage{graphicx}

\DeclareMathOperator*{\argmin}{argmin}
\DeclareMathOperator*{\minimize}{minimize}
\DeclareMathOperator*{\maximize}{maximize}
\DeclareMathOperator*{\subject}{subject \ to}

\newcommand{\dboost}{\textit{dboost}}

\newtheorem{theorem}{Theorem}
\newtheorem{cor}{Corollary}
\newtheorem{prop}{Proposition}
\newtheorem{defn}{Definition}

\DeclareMathOperator{\bone}{\bf 1}
\DeclareMathOperator{\bzero}{\bf 0}
\DeclareMathOperator{\bA}{\bf A}

\DeclareMathOperator{\blb}{\bf b}
\DeclareMathOperator{\bc}{\bf c}

\DeclareMathOperator{\bd}{\bf d}

\DeclareMathOperator{\bG}{\bf G}
\DeclareMathOperator{\bH}{\bf H}
\DeclareMathOperator{\bI}{\bf I}

\DeclareMathOperator{\bL}{\bf L}

\DeclareMathOperator{\bM}{\bf M}
\DeclareMathOperator{\bP}{\bf P}

\DeclareMathOperator{\bq}{\bf q}

\DeclareMathOperator{\br}{\bf r}

\DeclareMathOperator{\bs}{\bf s}

\DeclareMathOperator{\bu}{\bf u}
\DeclareMathOperator{\bV}{\bf V}
\DeclareMathOperator{\bv}{\bf v}

\DeclareMathOperator{\bw}{\bf w}

\DeclareMathOperator{\bx}{\bf x}

\DeclareMathOperator{\by}{\bf y}
\DeclareMathOperator{\bz}{\bf z}

\DeclareMathOperator{\balpha}{\bm \alpha}

\DeclareMathOperator{\btheta}{\bm \theta}

\DeclareMathOperator{\bXi}{\bm \Xi}

\DeclareMathOperator{\bzeta}{\bm \zeta}

\providecommand{\keywords}[1]{\textbf{\textit{Keywords:}} #1}

\newcommand{\toi}[2][i]{%
  \mathop{
    \mathrm{#2}^{( #1 )}
  }
}

\begin{document}

\title{Gradient boosting for convex cone predict and optimize problems}
\author{Andrew Butler and Roy H. Kwon \\ University of Toronto\\Department of Mechanical and Industrial Engineering}
\maketitle

\begin{abstract}
Prediction models are typically optimized independently from decision optimization. A `smart predict then optimize' (SPO) framework optimizes prediction models to minimize downstream decision regret. In this paper we present \textit{dboost}, the first general purpose implementation of smart gradient boosting for `predict, then optimize' problems. The framework supports convex quadratic cone programming and gradient boosting is performed by implicit differentiation of a custom fixed-point mapping. Experiments comparing with state-of-the-art SPO methods show that \textit{dboost} can further reduce out-of-sample decision regret.
\end{abstract}


\maketitle


\section{Introduction} \label{sec:dboost_intro}

Recently there has been a growing body of research on decision-aware predictive modelling (see for example \citep{Bert2019, Bert2020,Elma2020, Elma2020b, Grigas2021,Mandi2020,Wilder2019}). A traditional `predict, then optimize' framework treats the prediction estimation and decision optimization problem independently. As such, an  `objective mismatch' \citep{Lambert2020} can occur whereby improved prediction accuracy does not result in improved decision accuracy.

Conversely, the smart `predict, then optimize' (SPO) \citep{Elma2020}  framework optimizes prediction models in order to minimize the final downstream decision regret. To date, the SPO framework has been studied in a general setting for linear and decision tree regression models \citep{Elma2020, Elma2020b}. In this paper we present \dboost, a general purpose framework that combines the strength of gradient boosting with the SPO framework. Previous work \citep{Kon2021} considers gradient boosting for integrated prediction and optimization problems but only considers a small subset of optimization problems with linear inequality constraints. In contrast, the \dboost~framework is capable of supporting any optimization problem that can be cast as a convex quadratic cone program (QCP); and thus supports  linear \footnote{We refer the reader to the Supplementary Material for a discussion on the limitations of strict lower-level linear programming.}, quadratic and second-order cone programming with general convex cone constraints. We present a novel fixed-point implicit differentiation algorithm for computing the gradient of the SPO loss with respect to all of the cone program variables. The \dboost~framework is provided as an open-source Python package, available here: $\text{https://github.com/ipo-lab/dboost\_py}.$

The remainder of the paper is outlined as follows. We begin with a brief overview of related work on integrated prediction and optimization. In Section \ref{sec:dboost_method} we  present convex quadratic cone programming in the context of the SPO framework and provide the fixed-point implicit differentiation algorithm. We then present the \dboost~framework as a general extension of the gradient boosting algorithm proposed by \citet{Friedman2001}.  Experimental results are provided in Section \ref{sec:results} and demonstrate  that training prediction models with \dboost~can reduce decision regret by anywhere from $15\%-90\%$ in comparison to existing solutions.


\subsection{Related work} \label{sec:boost_work}
The \dboost~framework applies gradient boosting \citep{Friedman2001} to the SPO loss function \citep{Elma2020}, described in more detail in Section \ref{sec:dboost_method}. Optimizing the SPO loss by gradient descent methods is challenging as it requires computing the gradient of the optimal solution with respect to predicted costs. Local gradient based methods \citep{Mandi2020,Tan2020} and convex approximations  \citep{Butler2020IPO,Elma2020} have proven to be effective in comparison to a traditional `predict, then optimize' approach and growing empirical evidence supports a fully integrated estimation approach  (see for example \citep{Amos2019, Donti2017, Mandi2020, Tan2020, Wilder2019}). 

Related, \citet{Amos2017} provide a framework for learning \textit{linear} constrained QPs in an end-to-end trainable neural network. Backpropagation is performed by implicit differentiation of the KKT optimality conditions. Similarly,  \citet{Agrawal2020} present a differentiable optimization layer for \textit{linear} convex cone programs and compute gradients by implicit differentiation of the residual map of the conic homogeneous self-dual embedding. In contrast, we present an alternative differentiation technique, customized to the QCP program, that implicitly differentiates a fixed-point mapping of the Douglas-Rachford splitting iterations \citep{DR56}.

Most relevant is the work of \citet{Elma2020b} who present SPO trees (SPOT) for optimizing regression trees to minimize downstream decision regret. The authors also consider a random forest \citep{Breiman2001} implementation, but do not consider a gradient boosting ensemble approach.  \citet{Kon2021} consider  gradient boosting problems under the SPO framework for a subset of optimization problems;  namely with linear inequality constraints, but do not consider more general convex optimization problems.  To our knowledge,  \dboost~is the first `smart' gradient boosting implementation that supports a more general class of convex quadratic cone programs.


\section{Methodology} \label{sec:dboost_method}
We consider convex quadratic cone programs (QCP) with the following primal-dual form \citep{Odono2020}:
\begin{align} \label{eq:qcp}
\minimize \quad & \frac{1}{2} \bz^T\bP \bz + \bc^T \bz &  \maximize \quad  & -\frac{1}{2} \bz^T\bP \bz -  \blb^T\by \nonumber \\
\subject \quad  &  \bA \bz + \bs = \blb  & \subject \quad   &  \bP\bz +  \bA^T \by + \bc = 0 \nonumber \\
 & \bs \in \mathcal{K} \qquad &  & \by \in \mathcal{K}^*
\end{align}
where $\bz \in \mathbb{R}^{d_z}$,  $\by \in \mathbb{R}^{d_y}$ and $\bs \in \mathbb{R}^{d_y}$ denote the primal, dual and slack variables, respectively. The objective input variables are a symmetric positive semi-definite matrix $\bP \in  \mathbb{R}^{d_z \times d_z}$ and cost vector  $\bc \in \mathbb{R}^{d_z}$. The feasible region is defined by the matrix $\bA \in \mathbb{R}^{d_y \times d_z}$, the vector $\blb \in \mathbb{R}^{d_y}$ and  the nonempty convex cone $\mathcal{K}$ with associated dual cone $\mathcal{K}^*$. We denote the optimal solution to Program  $\eqref{eq:qcp}$  as $\bzeta^* = (\bz^*,\by^*, \bs^*) \in \mathbb{S} \subset \mathbb{R}^{d_z}\times  \mathcal{K} \times  \mathcal{K}^*$ with feasible region: $\mathbb{S} = \{ \bzeta \in \mathbb{R}^{d_z}\times  \mathcal{K} \times  \mathcal{K}^* \mid \bA \bz + \bs = \blb, \bP\bz +  \bA^T \by + \bc = 0 \}.$

In practice, the true cost, $\bc$, is not directly observable at decision time, and rather an associated feature vector $\bx \in  \mathbb{R}^{d_x}$ is observed. Given a training dataset  $ D = \{(  \toi{\bx},  \toi{\bc})\}_{i=1}^m$ we seek to estimate a prediction model $f \colon \mathbb{R}^{d_x} \times \mathbb{R}^{d_\theta } \to \mathbb{R}^{d_z}$ such that: $\toi{\hat{\bc}} =  f(\toi{\bx}, \btheta)$. We consider `additive' prediction models of the form: $f(\bx, \btheta) = \sum_{n = 0}^N \beta_n h(\bx, \balpha_n)$, with parameter  ${\btheta = \{( \beta_n, \balpha_n )\}_{n=1}^N}$ and  non-negative weight $\beta_n \geq 0 \ \forall n.$   In particular, we focus on the case where the functions, $h(\bx, \balpha_n)$, are regression trees and therefore the parameters $\balpha_n$ encode the feature component and splitting information.

 A traditional `predict, then optimize' approach would estimate $\btheta$ by minimizing a prediction loss (such as least-squares) and then `plug-in' the estimate $\toi{\hat{\bc}}$ to Program $\eqref{eq:qcp}$ in order to retrieve the `optimal' decision. In contrast, a `smart predict, then optimize' (SPO) approach estimates the prediction model parameters by minimizing the decision regret:

\begin{equation}\label{eq:spo_loss}
\begin{split}
    \ell_{\text{QSPO}} (\hat{\bc},\bc) = & \frac{1}{2} \bz^*( \hat{\bc}) ^T \bP \bz^*( \hat{\bc} ) + \bc^T \bz^*( \hat{\bc} )  - \frac{1}{2} \bz^*( \bc) ^T \bP \bz^*(\bc ) - \bc^T\bz^*(\bc ).
 \end{split}
\end{equation}
The SPO estimation process can be posed as a bi-level optimization:
\begin{equation} \label{eq:spo_bi_level}
 \begin{split}
 \minimize_{\btheta} \quad & \frac{1}{m} \sum_{i = 1}^m  \ell_{\text{QSPO}}(\toi{\hat{\bc}},  \toi{\bc})   \\
\subject \quad  &     \bz^{*}(\toi{\hat{\bc}})  =  \argmin_{\bz \in \mathbb{S}_z } \frac{1}{2} \bz^T\bP \bz  + \toi{\hat{\bc}}^T\bz.
 \end{split}
 \end{equation}
 We approximate a local solution to Program $\eqref{eq:spo_bi_level}$ by applying functional gradient descent  \citep{Friedman2001}. Specifically, at each iteration of gradient descent we must first solve $m$ QCPs to optimality and the gradient, $\nabla_{\toi{\hat{\bc}}}   \ell_{\text{QSPO}}$, is obtained by specialized $\argmin$ differentiation described below.

\subsection{Fixed-point argmin differentiation}\label{sec:dboost_method_fp}
Program $\eqref{eq:qcp}$ is solved by applying a Douglas-Rachford splitting to a homogeneous embedding of the QCP as described in \citet{Odono2020}. Specifically, we define the convex cone $\mathcal{C} = \mathbb{R}^{d_z} \times  \mathcal{K}^* $ and denote:

\begin{equation}
\bu = \begin{bmatrix}
\bz\\
\by
\end{bmatrix},
 \quad
\bv = \begin{bmatrix}
\bzero \\
\bs
\end{bmatrix},
 \quad
\bM = \begin{bmatrix}
\bP + \bI_{\bz} &   \bA^T \\
-\bA & \bzero
\end{bmatrix},
 \quad
\bq = \begin{bmatrix}
\bc\\
\blb
\end{bmatrix},
\end{equation}
where $ \bI_{\bz}$ denotes the identity matrix of same dimension as $\bz$. \citet{Odono2020} demonstrates that a direct application of operator splitting produces the following procedure; from any initial $\bu^0$ and $\bv^0$ then the following iterations converge to the optimal $\bzeta^*$ (if it exists):

\begin{subequations} \label{eq:dr}
\begin{align}
\tilde{\bu}^{k+1} & = (\bI_{\bu} + \bM)^{-1}(\bu^{k} + \bv^{k} - \bq) \label{eq:dr_1} \\
\bu^{k+1} & = \Pi_{\mathcal{C}}( \tilde{\bu}^{k+1} - \bv^{k}) \label{eq:dr_2}\\
\bv^{k+1} & = \bv^k + \bu^{k+1} - \tilde{\bu}^{k+1}, \label{eq:dr_3}
\end{align}
\end{subequations}
where $\Pi_{\mathcal{C}}$ denotes the Euclidean projection operator onto the set $\mathcal{C}$.

We recast the iterative procedure $\eqref{eq:dr}$ as a fixed-point iteration over variable $\bw^k = \tilde{\bu}^{k+1} - \bv^k$ and apply the {\textit{implicit function theorem}} \citep{Dontchev2009} to compute the required gradients. We begin with the following proposition. Note that  all proofs are available in the Supplementary Material report.

\begin{prop}\label{prop:dr_fixed_point}
Let $\bw^k = \tilde{\bu}^{k+1} - \bv^k$  and define $F \colon \mathbb{R}^{d_z + d_y} \times \mathbb{R}^{d_\theta} \to \mathbb{R}^{d_z + d_y}$. Then the iterations in Equation $\eqref{eq:dr}$  can be cast as a fixed-point iteration of the form $F(\bw,\btheta) = \bw$ given by:

\begin{equation}
\bw^{k+1} = (\bI_{\bw} + \bM)^{-1}(2\Pi_{\mathcal{C}}(\bw^k) - \bw^k -\bq) + \bw^k - \Pi_{\mathcal{C}}(\bw^k).
\end{equation}

\end{prop}
We denote the derivative of the projection operator as $D\Pi_{\mathcal{C}}$ and for ease of notation we drop the index $k$. The Jacobian, $\nabla_{\bw} F$, is therefore defined as:
\begin{equation}\label{eq:dr_jacob}
\nabla_{\bw} F = (\bI_{\bw} + \bM)^{-1}(2D\Pi_{\mathcal{C}}(\bw) - \bI_{\bw} ) + \bI_{\bw}- D\Pi_{\mathcal{C}}(\bw).
\end{equation}
Direct application of the implicit function theorem gives the desired Jacboian, $\nabla_{\btheta} \bw^*(\btheta)$, with respect to the parameter $\btheta$ at optimality $\bw^*$:
\begin{equation}\label{eq:jacob_v}
\nabla_{\btheta} \bw^*(\btheta) = [\bI_{\bw} - \nabla_{\bw} F(\bw^*(\btheta),\btheta) ]^{-1} \nabla_{\btheta} F(\bw^*(\btheta),\btheta).
\end{equation}
From the definition of $\bw^*$ we have that ${\bu^* =  \Pi_{\mathcal{C}}(\bw^*)}$ and therefore $\nabla_{\bw^*} \bu^* = D\Pi_{\mathcal{C}}(\bw^*)$.

Finally, learning $\btheta$ by gradient boosting ultimately requires computing the gradient $\partial \ell/\partial \bc$. From the chain rule we have $\partial \ell/\partial \bc =  \partial \ell/\partial \bz^* \cdot \partial \bz^*/\partial \bc $. In practice, however, it is computationally inefficient to form the Jacobian, $ \partial \bz^*/\partial \bc$, directly and instead we compute the action of the  left  matrix-vector  product  of  the  Jacobian  with  respect to the relevant  gradient, $\partial \ell/\partial \bz^*$, by solving the following system of equations. 

\begin{prop}\label{prop:dr_grads}
Denote the matrix $\bG=  (\bI_{\bw} + \bM)D\Pi_{\mathcal{C}}(\bw^*) +  \bI_{\bw} - 2D\Pi_{\mathcal{C}}(\bw^*)$ and define $\hat{ \bd }_{\bz}$ and $\hat{ \bd }_{\by}$ as:
\begin{equation}\label{eq:grads_dr}
\begin{split}
\begin{bmatrix}
\hat{ \bd }_{\bz}  \\
\hat{ \bd }_{\by}
\end{bmatrix}
& = \bG^{-T}D\Pi_{\mathcal{C}}(\bw^*)^T \begin{bmatrix}
-(\frac{\partial \ell }{\partial \bz^*})^T  \\
\bzero
\end{bmatrix}
\end{split}
\end{equation}
Then the gradients of the loss function, $\ell$, with respect to problem variables $\bP$, $\bc$, $\bA$ and $\blb$ are given by:
\begin{equation}\label{eq:dr_partials}
\begin{aligned}
\frac{\partial \ell   }{\partial \bP} & = \frac{1}{2} \Big(\hat{ \bd }_{\bz}   \bz^{*T} + \bz^* \hat{ \bd }_{\bz}^T \Big) & \qquad \frac{\partial \ell   }{\partial \bc} & = \hat{ \bd }_{\bz} & \qquad
 \frac{\partial \ell   }{\partial \bA} & = \by^* \hat{ \bd }_{\bz} ^T -  \hat{ \bd }_{\by}  \bz^{*T}    & \qquad \frac{\partial \ell   }{\partial \blb} & = \hat{ \bd }_{\by}\\
\end{aligned}
\end{equation}
\end{prop}



We now present the \dboost~algorithm which applies gradient boosting to minimize decision regret. We highlight two important differences between Algorithm \ref{algo:dboost} and the standard gradient boosting algorithm. First, on line $8$ we compute the gradient of the decision regret, $\ell_{\text{QSPO}}$, with respect to the estimated costs by first solving the optimal decisions, $\{\bz^*( \toi{\hat{\bc}} )\}_{i=1}^m$, and then performing implicit differentiation as described above. Similarly, the line search on line $10$ is with respect to the decision regret, $\ell_{\text{QSPO}}$, and therefore we must solve  $\{\bz^*( \toi{\hat{\bc}} + \beta h(\toi{\bx}, \balpha_n) )\}_{i=1}^m$ at each candidate value $\beta$. These differences make Algorithm \ref{algo:dboost} several orders of magnitude more computationally demanding than the typical gradient boosting algorithm. We refer the reader to the Supplementary Material report for a detailed discussion on computational efficiency.

\begin{algorithm}[H]
\caption{gradient boosting for min $ \ell_{\text{QSPO}}$ :}
\label{algo:dboost}
\begin{algorithmic}[1]
\Procedure{dboost}{}
    \State $f_0(\bx,\btheta) = \argmin_{\btheta} \text{Program} \eqref{eq:spo_bi_level}$
    \State Set $n = 0 $
    \State Set $0 <\epsilon_\beta  < 1$ and $0 <\epsilon_\ell  < 1$
    \While{ run = TRUE }
         \State $n = n+1$
	\State $\toi{\hat{\bc}} = f_{n-1}(\toi{\bx}, \btheta), i = 1, 2, ...,m$
        \State $\toi{\tilde{\bc}} = - \frac{\partial \ell_{\text{QSPO}}   }{\partial \toi{\hat{\bc}}}, i = 1, 2, ...,m$
        \State $\balpha_n =  \argmin_{\balpha} \sum_{i = 1}^m ( \toi{\tilde{\bc}}  - h(\toi{\bx}, \balpha))^2$
        \State $ \beta_n = \argmin_{\beta} \sum_{i = 1}^m \ell_{\text{QSPO}}(\toi{\hat{\bc}} + \beta h(\toi{\bx}, \balpha_n),  \toi{\bc}) $
        \State $f_{n}(\bx, \btheta) = f_{n-1}(\toi{\bx}, \btheta) +  \beta_n h(\toi{\bx}, \balpha_n)$
        \State $\ell_n = \sum_{i = 1}^m \ell_{\text{QSPO}}( f_n(\toi{\bx,\btheta)},  \toi{\bc})$
        \State $\Delta \ell = ( \ell_n - \ell_{n-1})/ \ell_{n-1}$
        \If{$\beta_n < \epsilon_\beta$ or $\Delta \ell < \epsilon_\ell$ or n = N  }
        \State run = FALSE
        \EndIf

    \EndWhile
    \State {\textbf{end}}

\EndProcedure

\end{algorithmic}
\end{algorithm}

The \dboost~algorithm focuses explicitly on learning the cost vector $\bc$. Indeed, in applications such as portfolio optimization, accurate cost vector estimation is shown to be an order of magnitude more important than estimation of $\bP$ \citep{Best1991, Chopra1993}.  From Proposition \ref{prop:dr_grads}, however, it is possible to generalize \dboost~in order to learn the other input variables: $\bP$, $\bA$ and $\blb$. For example, in optimal control it is often desirable to learn all input parameters jointly \citep{Amos2019}. Recent work in network flow optimization  \citep{Tan2020} and finance \citep{Butler2021Reg} advocate for data-driven approaches for learning regularization and constraint variables and is an interesting area of future research.


\section{Experiments}\label{sec:results}
We present three experiments comparing out-of-sample decision cost of \dboost~to  $5$ alternatives:
\begin{enumerate}
\item \textbf{CART:}  classification and regression tree optimized for prediction MSE \citep{Breiman2017}. 

\item \textbf{CART Forest:} A random forest \citep{Breiman2001} implementation of CART.

\item \textbf{SPOT:} SPO tree \citep{Elma2020b} optimized for SPO loss $\eqref{eq:spo_loss}$.

\item \textbf{SPOT Forest:} random forest implementation of  SPOT.

\item \textbf{MSE Boosting:} traditional gradient boosting \citep{Friedman2001} optimized for MSE.

\end{enumerate}
The CART and SPOT methods contain a single prediction tree whereas ensemble-based approaches use a maximum of $100$ trees. All methods consider maximum tree depths: $\{0,1,2 \}$ with a minimum split size of $50$ observations. Random forest implementations also consider an unlimited tree depth specification and perform bagging across feature variables and training data observations with a $50\%$ sampling rate. Experiments are evaluated over $10$ independent trials. Each trial consists of randomly generated training and out-of-sample datasets each with $m=1000$ observations. Performance is evaluated with respect to the total excess decision cost, given by:
\begin{equation}\label{eq:excess_cost}
\frac{  \sum_{i = 1}^m \ell_{\text{QSPO}}(\toi{\hat{\bc}},  \toi{\bc}) } {  \mid  \sum_{i = 1}^m  \frac{1}{2} \bz^*( \toi{\bc}) ^T \bP \bz^*(\toi{\bc} ) + \toi{\bc}^T\bz^*(\toi{\bc} ) \mid  }
\end{equation}
We consider three optimization problems: a noisy network-flow, a noisy quadratic program and a noisy portfolio optimization with noise levels: $\tau \in \{ 0.0, 0.5, 1.0 \}$. See the Supplementary Material report for implementation details.

\subsection{Results}
\begin{figure}[h]
  \centering
  \begin{subfigure}[b]{0.32\linewidth}
    \includegraphics[width=\linewidth ]{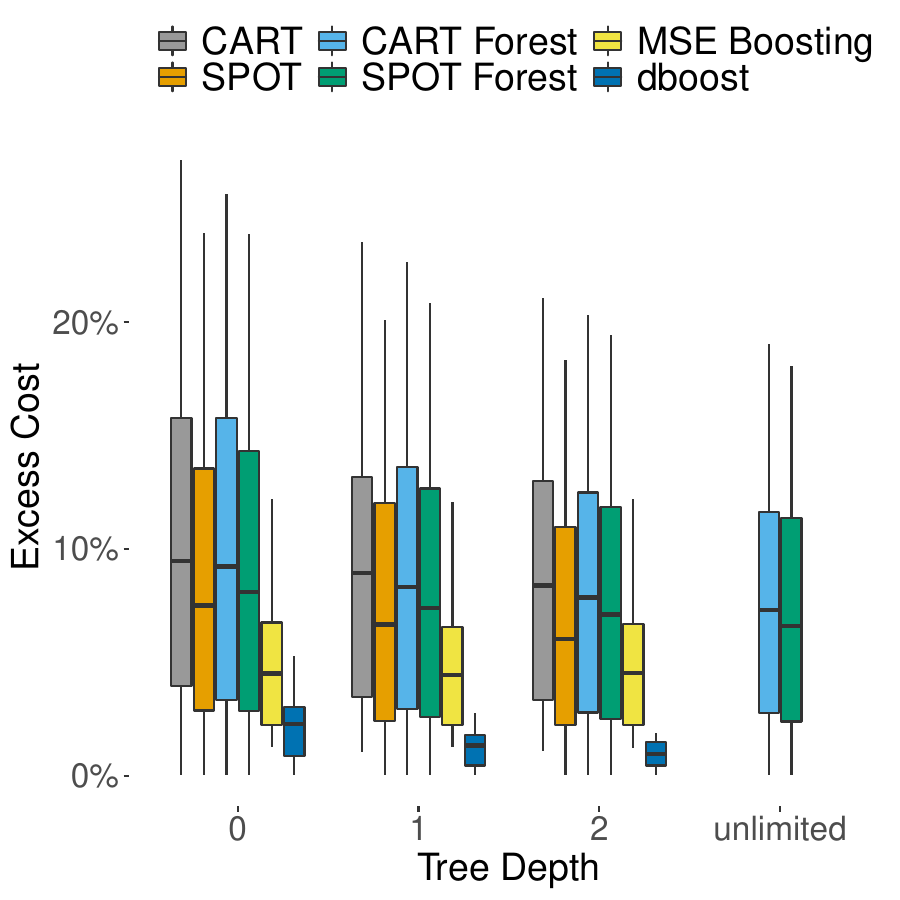}
    \caption{$\tau = 0.0$.}
  \end{subfigure}
  \begin{subfigure}[b]{0.32\linewidth}
    \includegraphics[width=\linewidth ]{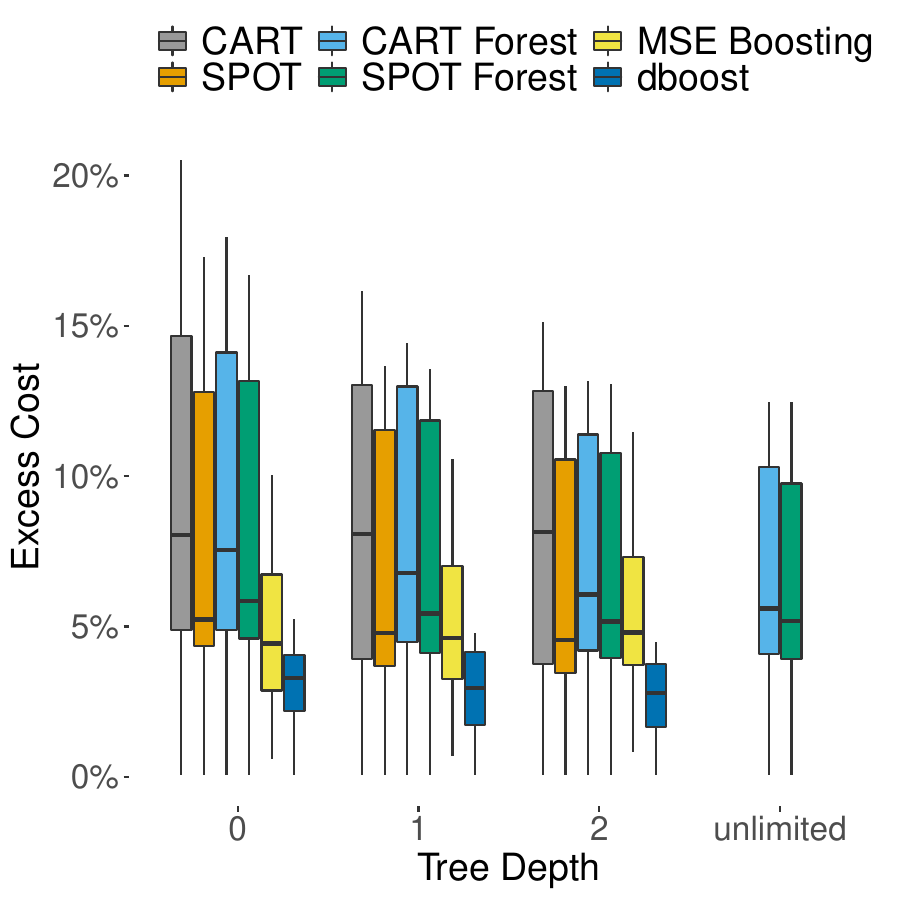}
    \caption{$\tau = 0.5$.}
  \end{subfigure}
    \begin{subfigure}[b]{0.32\linewidth}
    \includegraphics[width=\linewidth ]{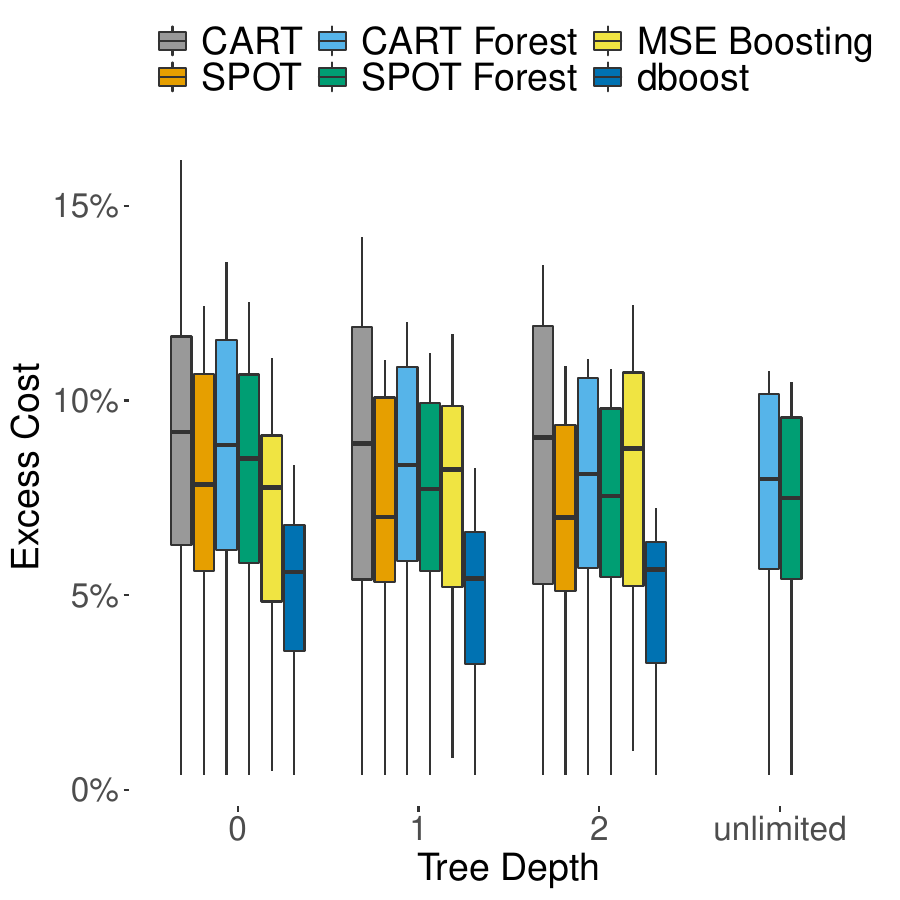}
    \caption{$\tau = 1.0$.}
  \end{subfigure}
  \caption{Out-of-sample excess cost for network flow problem with noise level $\tau \in \{0.0,0.5, 1.0\}$.}
  \label{fig:network}
\end{figure}

Figure \ref{fig:network} reports the out-of-sample excess cost for the noisy network flow problem.  Gradient boosting models produce excess costs that are on average $50\%$ lower than the corresponding CART and SPOT costs. In all cases the \dboost~model most effectively minimizes the decision cost, further reducing the excess cost of the MSE Boosting model by $30\%-60\%$, on average.  We observe that the reduction in cost is greatest when the noise level is low and in general deeper tree models tend to produce smaller costs. 

\begin{figure}[h]
  \centering
  \begin{subfigure}[b]{0.32\linewidth}
    \includegraphics[width=\linewidth ]{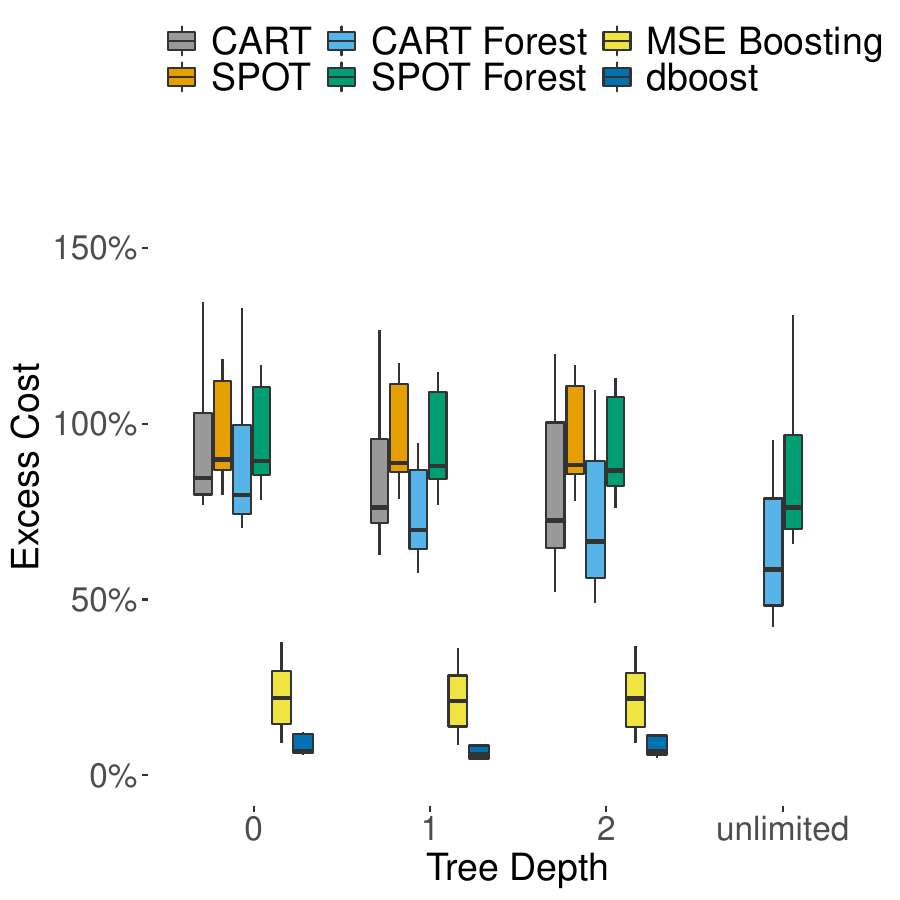}
    \caption{$\tau = 0$.}
  \end{subfigure}
  \begin{subfigure}[b]{0.32\linewidth}
    \includegraphics[width=\linewidth ]{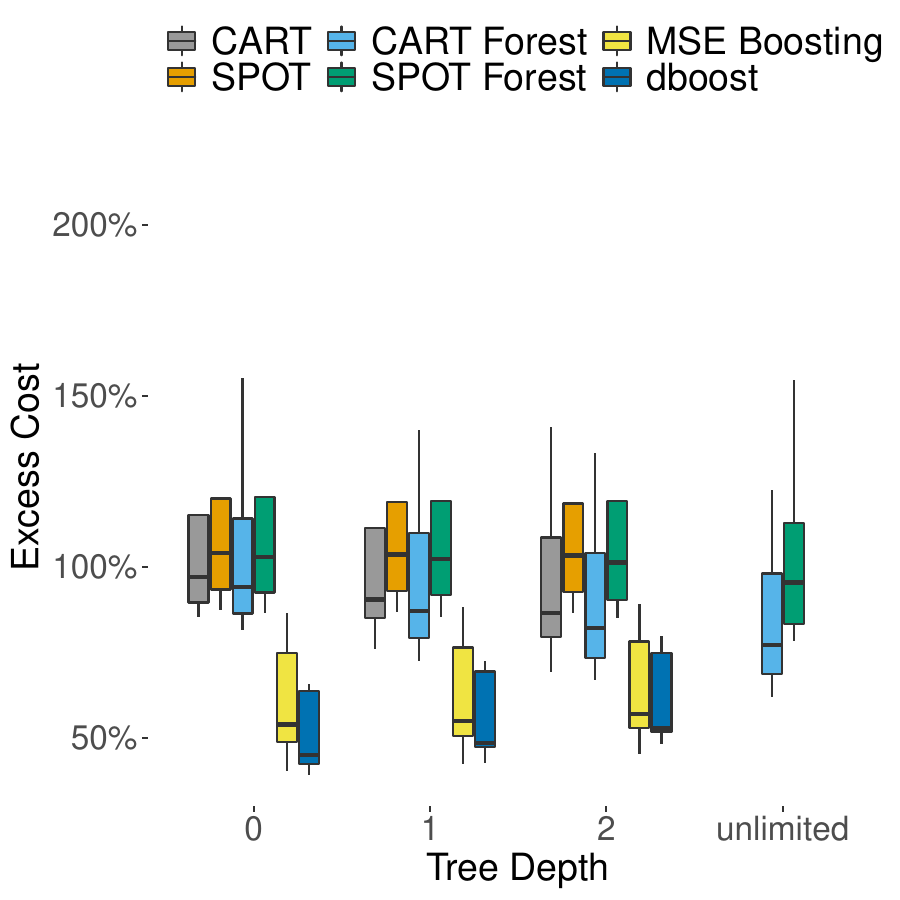}
    \caption{$\tau = 0.50$.}
  \end{subfigure}
    \begin{subfigure}[b]{0.32\linewidth}
    \includegraphics[width=\linewidth ]{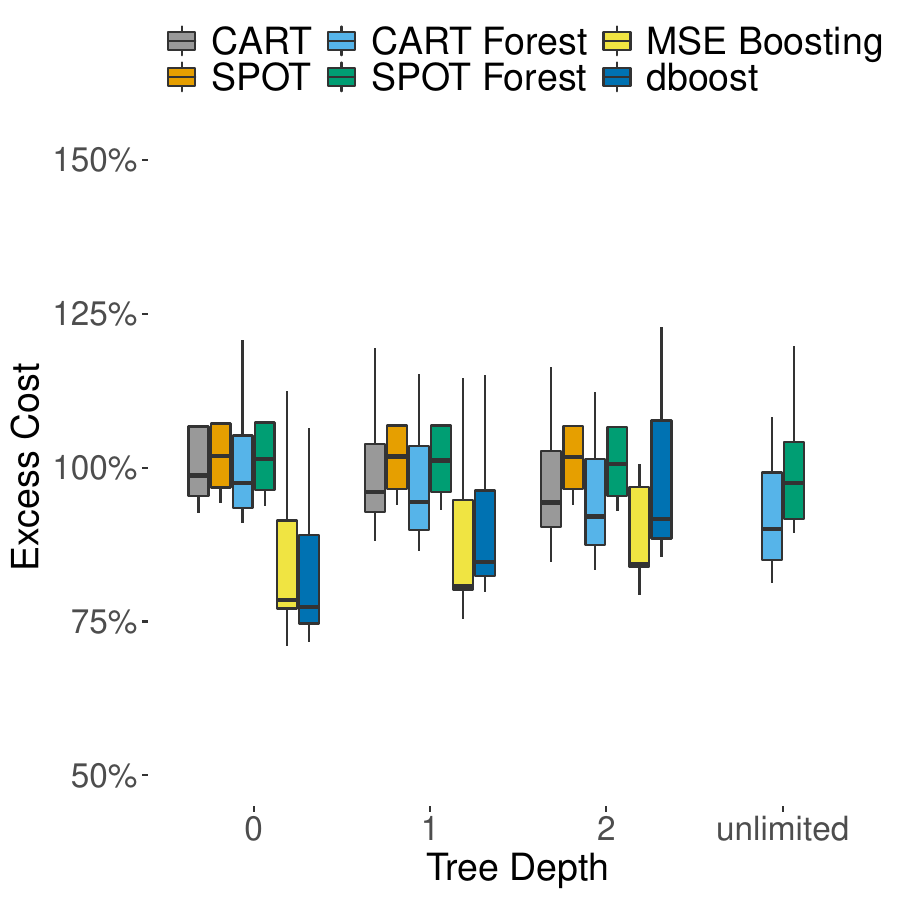}
    \caption{$\tau = 1.0$.}
  \end{subfigure}
 \caption{Out-of-sample excess cost for quadratic program problem with noise level $\tau \in \{0.0,0.5, 1.0\}$.}
  \label{fig:qp}
\end{figure}

Figure \ref{fig:qp} reports the out-of-sample excess cost for the noisy quadratic program problem.  Again, we observe that both gradient boosting models produce the smallest out-of-sample costs with excess costs that are $25\%-75\%$ smaller than the corresponding CART and SPOT costs. When $\tau = 0$, the average excess cost of the MSE Boosting model is approximately $22.2\%$ whereas \dboost~produces an average excess cost of approximately $5.8\%$; a further $75\%$ reduction in excess cost over the traditional gradient boosting model. In general the \dboost~model produces the smallest excess cost, however, as the noise level increases the difference in excess costs is less substantial. In fact, when $\tau = 1$ and tree depth is greater than zero, the MSE Boosting and unlimited CART Forest models produces marginally lower excess costs. Finally, observe that when the noise level is high ($\tau = 1$) both gradient boosted models with tree depth $0$ produce the smallest excess cost.

\begin{figure}[h]
  \centering
  \begin{subfigure}[b]{0.32\linewidth}
    \includegraphics[width=\linewidth ]{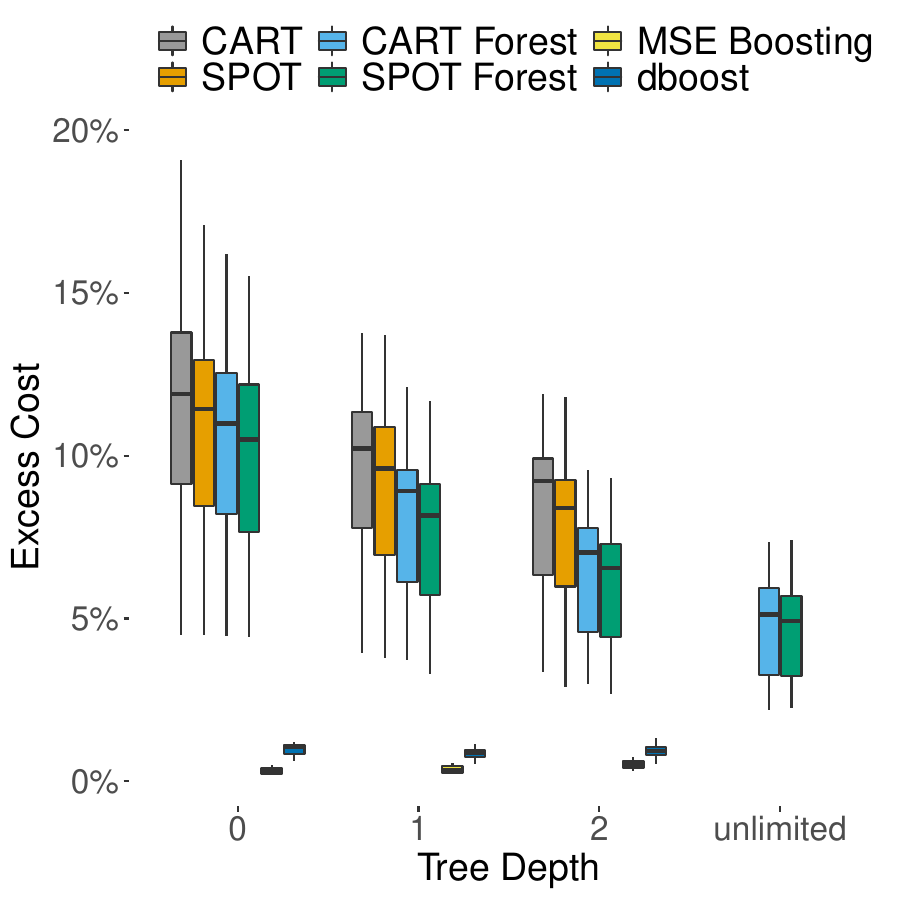}
    \caption{$\tau = 0$.}
  \end{subfigure}
  \begin{subfigure}[b]{0.32\linewidth}
    \includegraphics[width=\linewidth ]{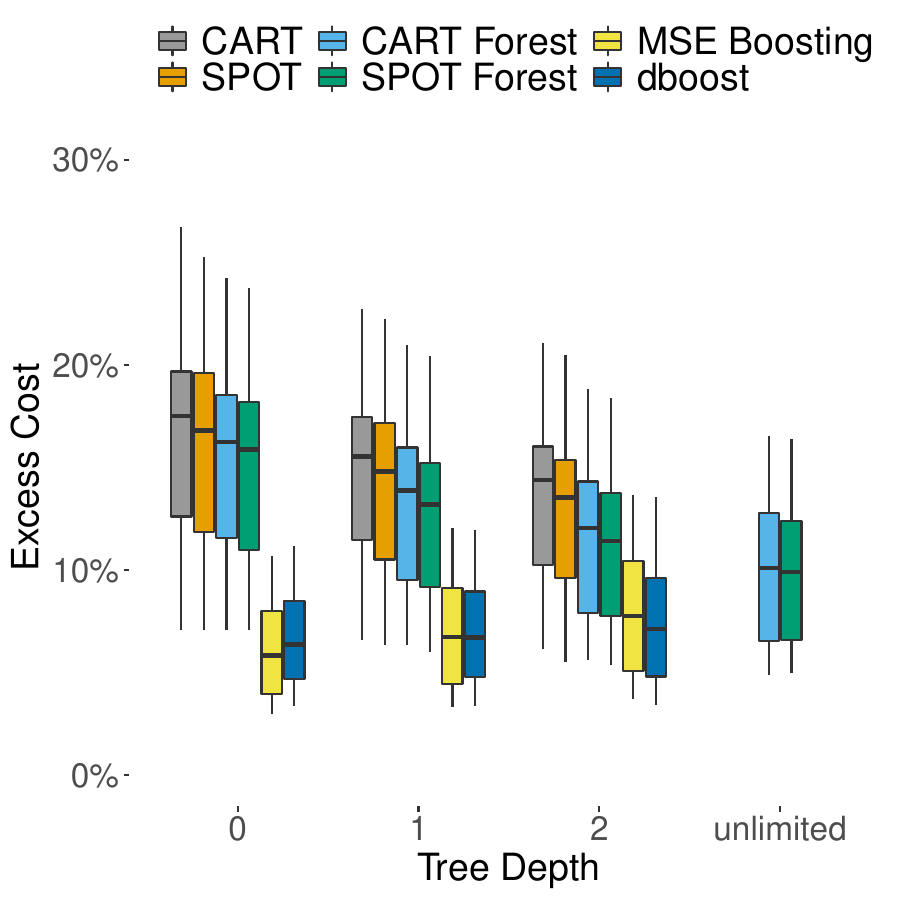}
    \caption{$\tau = 0.50$.}
  \end{subfigure}
    \begin{subfigure}[b]{0.32\linewidth}
    \includegraphics[width=\linewidth ]{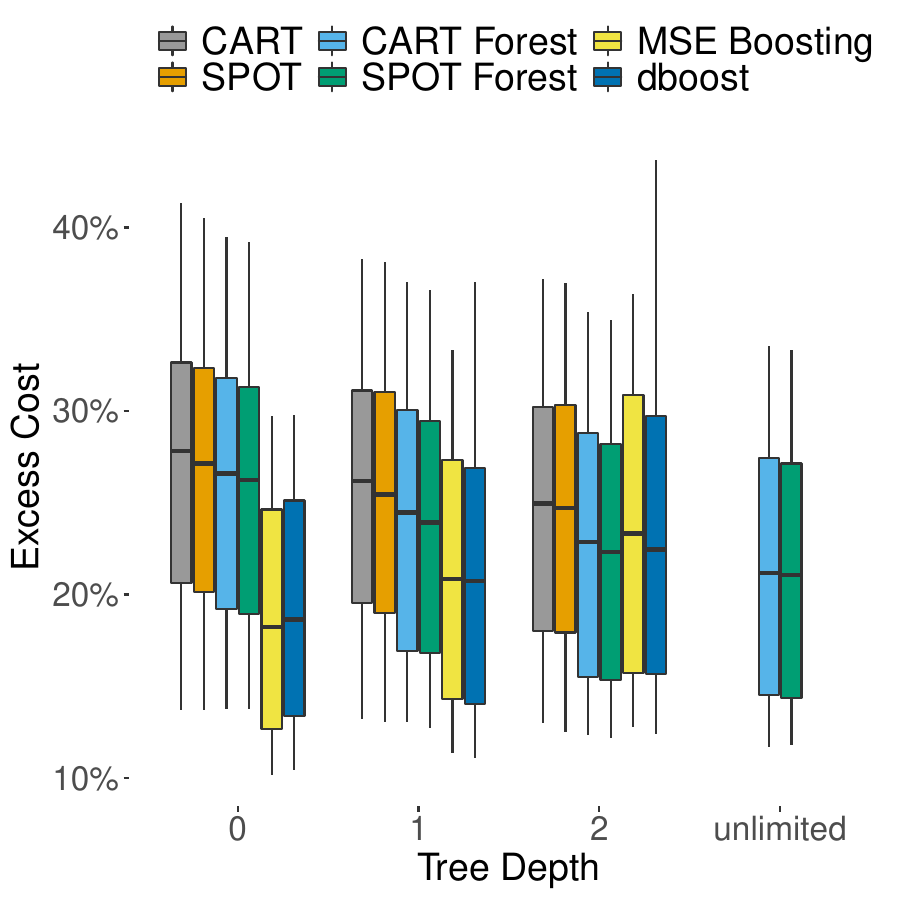}
    \caption{$\tau = 1.0$.}
  \end{subfigure}
 \caption{Out-of-sample excess cost for portfolio optimization problem with noise level $\tau \in \{0.0,0.5, 1.0\}$.}
  \label{fig:popt}
\end{figure}

Lastly, Figure \ref{fig:popt} reports the out-of-sample excess cost for the noisy portfolio optimization program problem. We observe that both gradient boosting models produce excess costs that are anywhere from $15\%-90\%$ smaller than the corresponding CART and SPOT costs.  In contrast to the prior experiments, in this experiment,  \dboost~provides no benefit over traditional gradient boosting and instead produces marginally higher costs on average. We observe that  as the noise level increases, MSE Boosting and \dboost~produce very similar excess costs. Across all noise levels, MSE Boosting with tree depth $0$ produces the smallest excess cost.

\subsection{Conclusion}
We presented \dboost, a general purpose framework for building `smart' gradient boosting prediction models. Experimental results should be interpreted as a proof-of-concept and the author's acknowledge that further testing is required in order to better determine the efficacy of a `smart' gradient boosting approach. Nonetheless the experimental results demonstrate that \dboost~is competitive with existing models and in many cases results in reduced out-of-sample decision regret. The \dboost~algorithm is significantly more computationally demanding than the typical gradient boosting algorithm. Improving the performance and scalability of \dboost~is an important area of future research.

\bibliographystyle{plainnat}
\bibliography{Bibliography}

\appendix
\section{Proof of Proposition $1$}
\citet{Odono2020} demonstrates that a direct application of operator splitting produces the following procedure; from any initial $\bu^0$ and $\bv^0$ then the following iterations converge to the optimal $\bzeta^*$ (if it exists):

\begin{subequations} \label{eq:dr}
\begin{align}
\tilde{\bu}^{k+1} & = (\bI_{\bu} + \bM)^{-1}(\bu^{k} + \bv^{k} - \bq) \label{eq:dr_1} \\
\bu^{k+1} & = \Pi_{\mathcal{C}}( \tilde{\bu}^{k+1} - \bv^{k}) \label{eq:dr_2}\\
\bv^{k+1} & = \bv^k + \bu^{k+1} - \tilde{\bu}^{k+1}, \label{eq:dr_3}
\end{align}
\end{subequations}
where $\Pi_{\mathcal{C}}$ denotes the Euclidean projection operator onto the set $\mathcal{C}$. The iterations in Equation $\eqref{eq:dr}$  can be cast as a fixed-point iteration as follows. Let $\bw^k = \tilde{\bu}^{k+1} - \bv^k$, then:

\begin{equation}\label{eq:app_u}
\bu^{k+1} =  \Pi_{\mathcal{C}}( \tilde{\bu}^{k+1} - \bv^{k})   = \Pi_{\mathcal{C}}( \bw^k ),
\end{equation}
and:
\begin{equation}\label{eq:app_v}
{\bv}^{k+1} =\bv^k + \bu^{k+1} - \tilde{\bu}^{k+1}=  \Pi_{\mathcal{C}}( \bw^k ) - \bw^k.
\end{equation}
Substituting Equations $\eqref{eq:app_u}$ and $\eqref{eq:app_v}$ into Equation $\eqref{eq:dr_1}$ gives the desired fixed-point iteration:
\begin{equation}
\begin{split}
\bw^{k+1} &  =  \tilde{\bu}^{k+2} - \bv^{k+1}\\
& =  (\bI_{\bu} + \bM)^{-1}(\bu^{k+1} + \bv^{k+1} - \bq)- \bv^{k+1}\\
& = (\bI_{\bw} + \bM)^{-1}(2\Pi_{\mathcal{C}}(\bw^k) - \bw^k -\bq) + \bw^k - \Pi_{\mathcal{C}}(\bw^k).
\end{split}
\end{equation}

\section{Proof of Proposition $2$}

We begin with the following definitions.
 \begin{defn}
Let $F \colon \mathbb{R}^{d_w} \times \mathbb{R}^{d_\theta} \to \mathbb{R}^{d_w}$ be a continuously differentiable function with variable $\bw$ and parameter $\btheta$. We define $\bw^*$ as a {\bf{fixed-point}} of $F$ at $(\bw^*,\btheta)$ if:
$$
F(\bw^*,\btheta) = \bw^*.
$$
\end{defn}

\begin{defn}
The {\bf{residual map}}, $G \colon \mathbb{R}^{d_w} \times \mathbb{R}^{d_\theta} \to \mathbb{R}^{d_w}$ of  a fixed point, $(\bw^*,\btheta)$, of $F$ is given by:
$$
G(\bw^*,\btheta) = F(\bw^*,\btheta) - \bw^* = 0.
$$
\end{defn}
The {\textbf{implicit function theorem}}, as defined by \citet{Dontchev2009},  then  provides the conditions on $G$ for which the Jacobian of the solution mapping with respect to $\btheta$ is well defined.

\begin{theorem}
Let $G \colon \mathbb{R}^{d_w} \times \mathbb{R}^{d_\theta} \to \mathbb{R}^{d_w}$ be a continuously differentiable function in a neighborhood of $(\bw^*, \btheta)$ such that $G(\bw^*, \btheta) = 0$. Denote the non-singular partial Jacobian of $G$ with respect to $\bw^*$ as $\nabla_{\bw^*} G(\bw^*, \btheta)$. Then $\bw^*(\btheta)$ is an implicit function of $\btheta$ and is continuously differentiable in a neighborhood, $\Theta$, of $\btheta$ with Jacobian:
\begin{equation}
\nabla_{\btheta} \bw^*(\btheta) = -[\nabla_{\bw^*} G(\bw^*(\btheta),\btheta) ]^{-1} \nabla_{\btheta} G(\bw^*(\btheta),\btheta) \quad \forall \quad \btheta \in \Theta.
\end{equation}

\end{theorem}

\begin{cor}\label{cor:if}
Let $F \colon \mathbb{R}^{d_w} \times \mathbb{R}^{d_\theta} \to \mathbb{R}^{d_w}$ be a continuously differentiable function with fixed-point $(\bw^*,\btheta)$. Then $\bw^*(\btheta)$ is an implicit function of $\btheta$ and is continuously differentiable in a neighborhood, $\Theta$, of $\btheta$ with Jacobian:
\begin{equation}
\nabla_{\btheta} \bw^*(\btheta) = [\bI_{\bw} - \nabla_{\bw^*} F(\bw^*(\btheta),\btheta) ]^{-1} \nabla_{\btheta} F(\bw^*(\btheta),\btheta) \quad \forall \quad \btheta \in \Theta.
\end{equation}
\end{cor}

We define $F \colon \mathbb{R}^{d_z + d_y} \times \mathbb{R}^{d_\theta} \to \mathbb{R}^{d_z + d_y}$ as:
\begin{equation}
F(\bw^*,\btheta) = (\bI_{\bw} + \bM)^{-1}(2\Pi_{\mathcal{C}}(\bw^*) - \bw^* -\bq) + \bw^* - \Pi_{\mathcal{C}}(\bw^*).
\end{equation}
The Jacobian, $\nabla_{\bw^*} F$, is therefore defined as:
\begin{equation}
\nabla_{\bw^*} F = (\bI_{\bw} + \bM)^{-1}(2D\Pi_{\mathcal{C}}(\bw^*) - \bI_{\bw} ) + \bI_{\bw}- D\Pi_{\mathcal{C}}(\bw^*).
\end{equation}
We perform the left multiplication of $F$ by $(\bI_{\bw} + \bM)$  and compute the partial derivative of $F$ with respect to all other problem variables at the fixed-point $\bw^*$ as follows:
\begin{equation}
\begin{split}
(\bI_{\bw} + \bM) \partial F(\bw^*)  & =  -\partial \bq + \partial \bM \bw^* - \partial \bM \Pi_{\mathcal{C}}(\bw^*) - \partial \bM F \\
 & =  -\partial \bq + \partial \bM ( \bw^* -  \Pi_{\mathcal{C}}(\bw^*) - F(\bw^*))\\
 &  = -\partial \bq  - \partial \bM (\bI_{\bw} + \bM)^{-1}(2\Pi_{\mathcal{C}}(\bw^*) - \bw^* -\bq)\\
 & =  -\partial \bq - \partial \bM \bu^*
\end{split}
\end{equation}
It follows then that $\partial F(\bw^*)$ is given by:
\begin{equation}\label{eq:app_partial_F_dr}
\partial F(\bw^*) = -\big[ \bI_{\bw} + \bM \big]^{-1}
\begin{bmatrix}
 \partial \bc + \frac{1}{2} (\partial \bP +  \partial \bP^T)\bz^* + \partial \bA^T \by^*\\
 \partial \blb - \partial \bA \bz^*
 \end{bmatrix}
\end{equation}
Applying Corollary \ref{cor:if}  we compute the gradient action of Equation $\eqref{eq:app_partial_F_dr}$ and the left matrix-vector product of the transposed Jacobian with the gradient, $\frac{\partial \ell }{\partial \bz^*}$, to arrive at the desired result.
\begin{equation}\label{eq:app_grads_dr}
\begin{split}
\begin{bmatrix}
\hat{ \bd }_{\bz}  \\
\hat{ \bd }_{\by}
\end{bmatrix}
& =
\big[ \bI_{\bw} + \bM \big]^{-T} \big[ D\Pi_{\mathcal{C}}(\bw^*) - (\bI_{\bw} + \bM)^{-1}(2D\Pi_{\mathcal{C}}(\bw^*) - \bI_{\bw} )   \big]^{-T}  D\Pi_{\mathcal{C}}(\bw^*)^T \begin{bmatrix}
-(\frac{\partial \ell }{\partial \bz^*})^T  \\
\bzero
\end{bmatrix}\\
& = \big[  (\bI_{\bw} + \bM)D\Pi_{\mathcal{C}}(\bw^*) - (2D\Pi_{\mathcal{C}}(\bw^*) - \bI_{\bw} ) \big]^{-T}D\Pi_{\mathcal{C}}(\bw^*)^T \begin{bmatrix}
-(\frac{\partial \ell }{\partial \bz^*})^T  \\
\bzero
\end{bmatrix}
\end{split}
\end{equation}
Finally, the gradients of the loss function, $\ell$, with respect to problem variables $\bP$, $\bc$, $\bA$ and $\blb$ are given by:
\begin{equation}
\begin{aligned}
\frac{\partial \ell   }{\partial \bP} & = \frac{1}{2} \Big(\hat{ \bd }_{\bz}   \bz^{*T} + \bz^* \hat{ \bd }_{\bz}^T \Big) & \qquad \frac{\partial \ell   }{\partial \bc} & = \hat{ \bd }_{\bz}  \\
\frac{\partial \ell   }{\partial \bA} & = \by^* \hat{ \bd }_{\bz} ^T -  \hat{ \bd }_{\by}  \bz^{*T}    & \qquad \frac{\partial \ell   }{\partial \blb} & = \hat{ \bd }_{\by}
\end{aligned}
\end{equation}

\section{Implementation details}
Experiments are conducted on an Apple Mac Pro computer (2.7 GHz 12-Core Intel Xeon E5,128 GB 1066 MHz DDR3 RAM) running macOS `Catalina'. All computations are run on an unloaded, single-threaded CPU. The software was written using the R programming language (version 4.0.0). 

Synthetic data is generated as follows. Feature variables, $\bx$, are generated by randomly drawing from the Uniform distribution $\mathcal{U}(-1,1)$. The cost values, $\bc$, are generated according to the polynomial model:
\begin{equation}\label{eq:c_poly}
\bc = \bH_0 + \sum_{j =1}^p \bH_j \bx^j + \tau \epsilon,
\end{equation}
with intercept $\bH_0$ and regression coefficients $\bH_j \in \mathbb{R}^{d_z \times d_x}$. The parameter $p$ controls the polynomial degree. Regression coefficients are sparse with each element of $\bH_j$ having a $50\%$ probability of being $0$ and $50\%$ probability of being nonzero $\mathcal{U}(-1,1)$. We let $\epsilon \sim \mathcal{N}(\bzero, \bone)$ and the scalar value $\tau$ controls the amount of noise in the data. All experiments consider three noise levels: $\tau \in \{ 0.0, 0.5, 1.0 \}$ and maximum tree depths: $\{0,1,2 \}$ with a minimum split size of $50$ observations. Random forest implementations also consider an unlimited tree depth specification and perform bagging across feature variables and training data observations with a $50\%$ sampling rate. We present three optimization problems: a noisy network-flow, a noisy quadratic program and a noisy portfolio optimization, described below.

\noindent \textbf{1. Noisy network-flow.} We consider a continuous network flow problem over a directed graph with $5$ nodes. Network edges are randomly generated with the probability that node $i$ flows to node $j$ given by $Pr(i \rightarrow j) = 0.75^{ | i-j-1 |}$. The number of decision variables, $d_z$, is determined by the number of edges.  Cost values are generated according to $\eqref{eq:c_poly}$ with  $ \bH_0 \sim \mathcal{N}(-\bone, \bone)$, $p = 3$ and $d_x = 5$. We consider a linear objective, $\ell_{\text{QSPO}} (\hat{\bc},\bc) =   \bc^T \bz^*( \hat{\bc} )  - \bc^T\bz^*(\bc )$, with a lower-level $L_2$-norm regularized network-flow optimization program:
\begin{equation}\label{eq:prob_lp}
\begin{split}
\minimize  & \quad  \hat{\bc}^T \bz + \frac{1}{2} \lVert  \bz \rVert_2^2   \\
\subject  & \quad  \bA \bz = \blb, \bzero \leq \bz \leq \bone .\\
\end{split}
\end{equation}
\textbf{2. Noisy quadratic program.}  We consider a quadratic objective, $$\ell_{\text{QSPO}} (\hat{\bc},\bc) =  \frac{1}{2} \bz^*( \hat{\bc}) ^T \bP \bz^*( \hat{\bc} ) + \bc^T \bz^*( \hat{\bc} )  - \frac{1}{2} \bz^*( \bc) ^T \bP \bz^*(\bc ) - \bc^T\bz^*(\bc )$$ with lower-level equality constrained quadratic program of the form:

\begin{equation}\label{eq:prob_qp}
\begin{split}
\minimize  & \quad  \hat{\bc}^T \bz + \frac{1}{2} \bz^T \hat{\bP} \bz   \\
\subject  & \quad  \bA \bz = \blb \\
\end{split}
\end{equation}
with number of decision variables $d_z = 25$ and randomly generated constraint matrix $\bA \in \mathbb{R}^{3 \times d_z}$ where $Pr(A_{jk} = 0) =Pr(A_{jk} = 1) = 0.50$. The vector $\blb$ is chosen to guarantee that the problem is non-empty.  Cost values are generated according to $\eqref{eq:c_poly}$ with  $ \bH_0  = \bzero $, $p = 3$ and $d_x = 5$. The positive definite matrix, $\bP$, is subject to estimation error: $\hat{\bP} = \bP + 0.1\bXi$ where $\bXi  = \frac{1}{n} \bL^T \bL$, $\bL \in \mathbb{R}^{10d_z \times d_z} $ and entries of $\bL  \sim \mathcal{N}(\bzero, \bone)$.  
\newline
\noindent \textbf{3. Noisy portfolio optimization.} We consider a linear objective, $\ell_{\text{QSPO}} (\hat{\bc},\bc) =   \bc^T \bz^*( \hat{\bc} )  - \bc^T\bz^*(\bc )$, with lower-level Markowitz \citep{Markowitz1952} mean-variance optimization program:
\begin{equation}\label{eq:prob_qp}
\begin{split}
\minimize  & \quad  \hat{\bc}^T \bz  \\
\subject  & \quad  \bone^T \bz = 1, \bz \geq 0 \\
& \sqrt{ \bz^T \bV \bz} \leq \sigma,
\end{split}
\end{equation}
where $\bz_j$ denotes the proportion of capital allocated to asset $j$. The cost values, $\bc$, are the negative asset returns and are generated according to $\eqref{eq:c_poly}$ with  $ \bH_0  = \sim \mathcal{N}(\bzero, \bone) $, $p = 3$ and $d_x = 5$. The covariance matrix, $\bV$, is generated according to a linear factor model $\bV  = \bL^T \bL + \varepsilon \bI_{\bz} $, with $\varepsilon = 0.01$ and factor matrix $\bL \in \mathbb{R}^{4 \times d_z}$  with entries $\bL  \sim \mathcal{U}(-1, 1)$. The second-order cone constraint limits the maximum risk of the portfolio. In each trial we set $\sigma = d_z^{-1} \sqrt{ \bone^T \bV \bone}$.

\section{Motivating example}
We are motivated by the work of \citep{Elma2020,Elma2020b} who demonstrate that optimizing prediction model parameters to minimize decision regret produces prediction models with lower complexity and improved out-of-sample performance. Below, we provide a motivating example to help illustrate the behaviour of \dboost~in comparison to a traditional gradient boosting model trained to minimize prediction mean-square error (MSE).  We seek to maximize the return (minimize the cost) of a portfolio of two assets subject to linear constraints:

\begin{equation} \label{eq:mot_ex}
 \begin{split}
 \minimize_{\bz} \quad & -\bz_1\br_1 - \bz_2\br_2 +\frac{1}{2} \lVert \bz \rVert_2^2   \\
\subject \quad  &      \bz_1 + \bz_2 = 1\\
\quad & \bz \geq 0
 \end{split}
 \end{equation}
 where $\bz_1$ and $\bz_2$ denote the proportion of capital allocated to asset $1$ and $2$, respectively. Here, the norm penalty $ \lVert \bz \rVert_2^2$ enforces portfolio diversification when differences in returns are small.  We generate a dataset of $500$  observations where the feature, $\bx$, is drawn from the standard uniform distribution and the return of each asset is given by $\br_1 = \bx + \epsilon$ and $\br_2 = \bx + \text{sin}(3\bx) + \epsilon$ where $\epsilon \sim \mathcal{N}(0, 0.10)$.

\begin{figure}[h]
  \centering
  \begin{subfigure}[b]{0.45\linewidth}
    \includegraphics[width=\linewidth ]{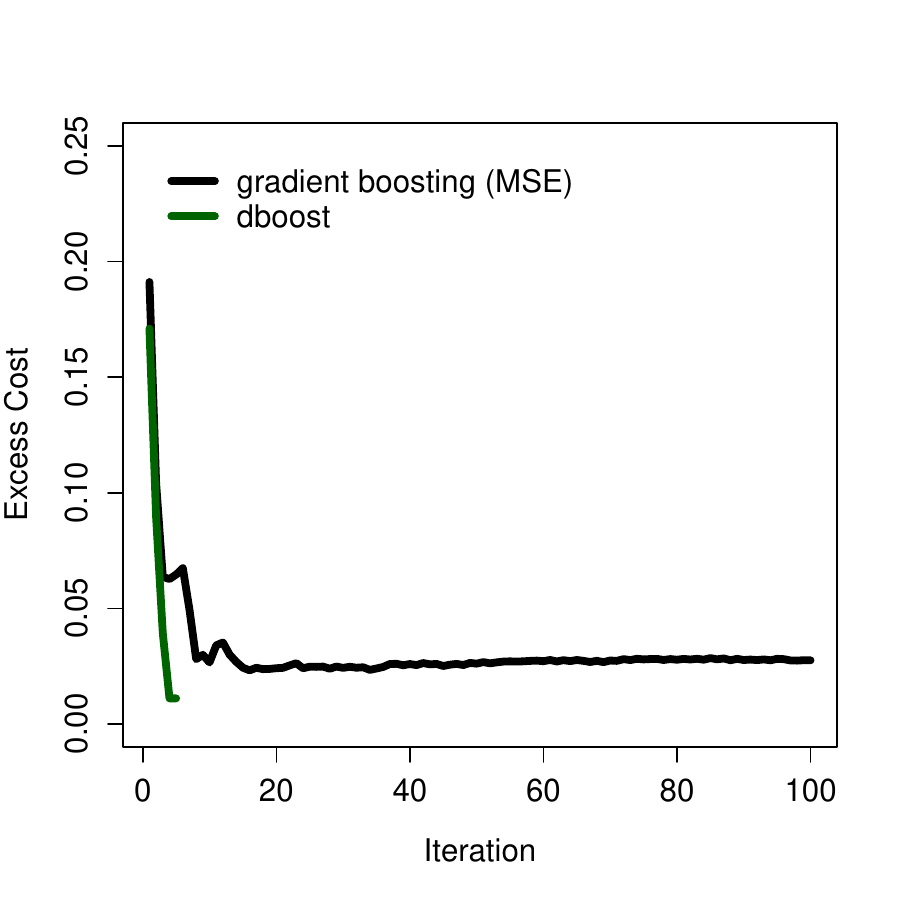}
    \caption{Excess decison cost}
    \label{fig:ex_1}
  \end{subfigure}
  \begin{subfigure}[b]{0.45\linewidth}
    \includegraphics[width=\linewidth ]{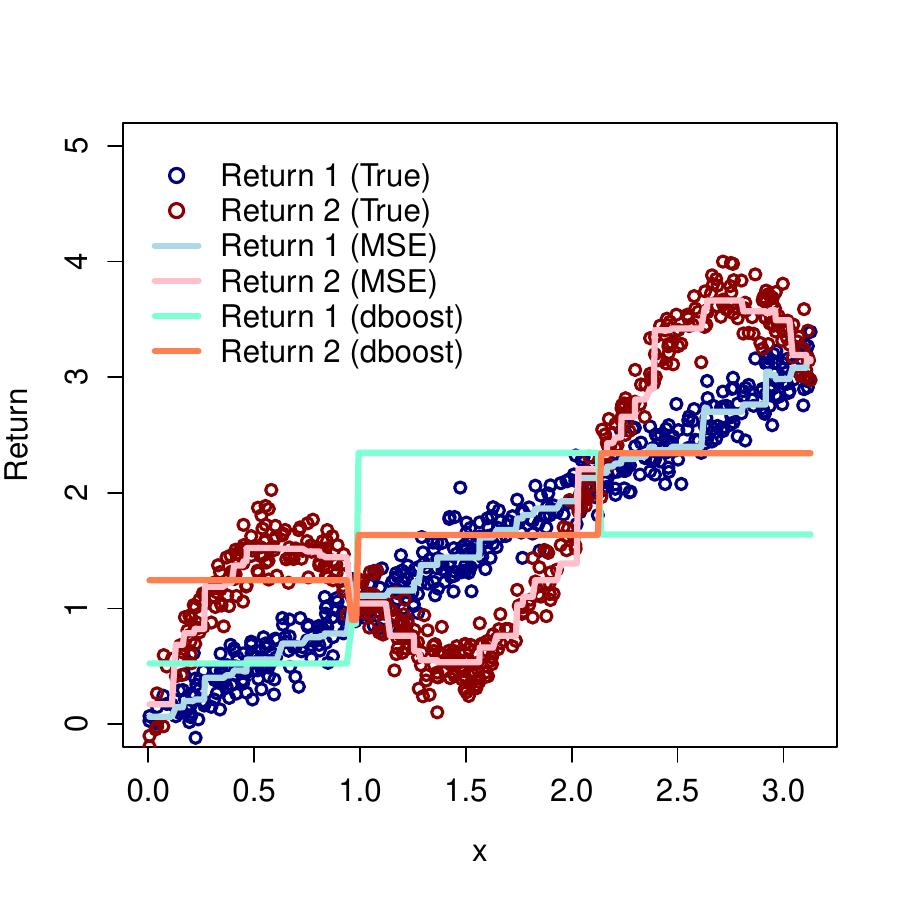}
    \caption{Return forecasts.}
    \label{fig:ex_2}
  \end{subfigure}
  \caption{In-sample excess decison cost and return forecasts for  MSE and \dboost~prediction models.}
  \label{fig:ex}
\end{figure}

Figure \ref{fig:ex_1} plots the total excess decision cost at each iteration of gradient boosting. Recall, that the traditional model is unaware as to how the predictions will ultimately be used in the context of the final downstream optimization model, and as such continues to add trees to the ensemble in an effort to minimize the prediction MSE. In contrast, the \dboost~model explicitly minimizes decision regret; the boosting algorithm terminates after $5$ iterations with {\textit{equal}} decision accuracy.  Figure \ref{fig:ex_2} plots the predictions model forecasts as a function of the feature variable $\bx$. In this example there are two decision boundaries: $\bx \approx 1$ and $\bx \approx 2$, where the optimal decision changes from favouring one asset over the other.  The \dboost~prediction model correctly identifies these approximate decision boundaries with substantially fewer trees and therefore avoids having to overfit the training dataset.

\section{Limitations}
A potential limitation of the SPO framework occurs when the lower-level program is strictly linear as the solution to the linear program may not be continuously differentiable with respect to $\hat{\bc}$ \citep{Wilder2019}. \citet{Elma2020} proposed replacing the SPO loss with a sub-differentiable convex surrogate loss (SPO$+$). Alternatively, in many practical settings it is sufficient to augment the lower-level program with an $L_2$-norm penalty \citep{Wilder2019}, or a log-barrier term and apply an early stopping criteria \citep{Mandi2020}.

\section{Computational efficiency}

We consider learning the cost vector for the noisy portfolio optimization problem. In general, we find that the computation time required to fit a \dboost~model scales linearly with the number of trees in the ensemble. Furthermore, the computation time is expected to scale linearly with the number of observations, $m$, in the training dataset. Indeed, the dominant computational effort in the \dboost~Algorithm  occurs on line $8$, where in order to compute the gradient of the decision regret, $\ell_{\text{QSPO}}$, with respect to the estimated costs, we must first solve for the optimal decisions, $\{\bz^*( \toi{\hat{\bc}} )\}_{i=1}^m$. Similarly, the line search on line $10$ is with respect to the decision regret, $\ell_{\text{QSPO}}$, and therefore we must solve for the optimal decisions  $\{\bz^*( \toi{\hat{\bc}} + \beta h(\toi{\bx}, \balpha_n) )\}_{i=1}^m$ at each candidate value $\beta$. As such, the dominant computational effort in the \dboost~algorithm is in repeatedly solving for $m$ optimal decisions at each iteration. 

\begin{figure}[H]
  \centering
  \begin{subfigure}[b]{0.32\linewidth}
    \includegraphics[width=\linewidth ]{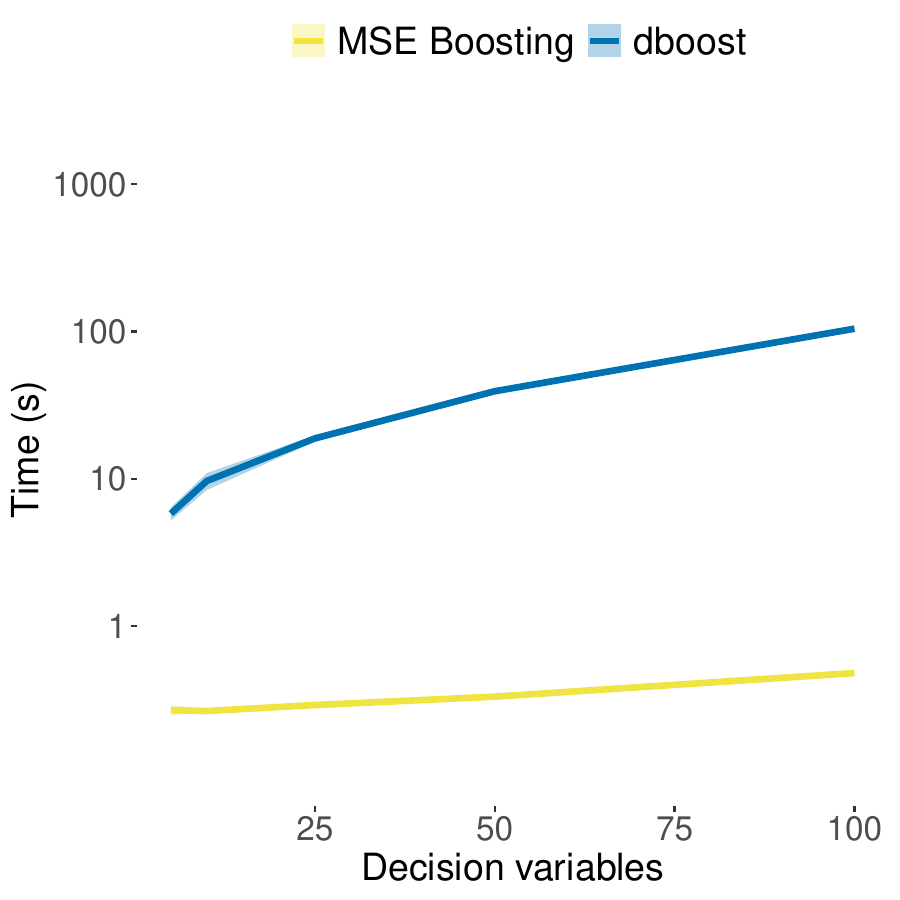}
    \caption{$m = 50$.}
  \end{subfigure}
  \begin{subfigure}[b]{0.32\linewidth}
    \includegraphics[width=\linewidth ]{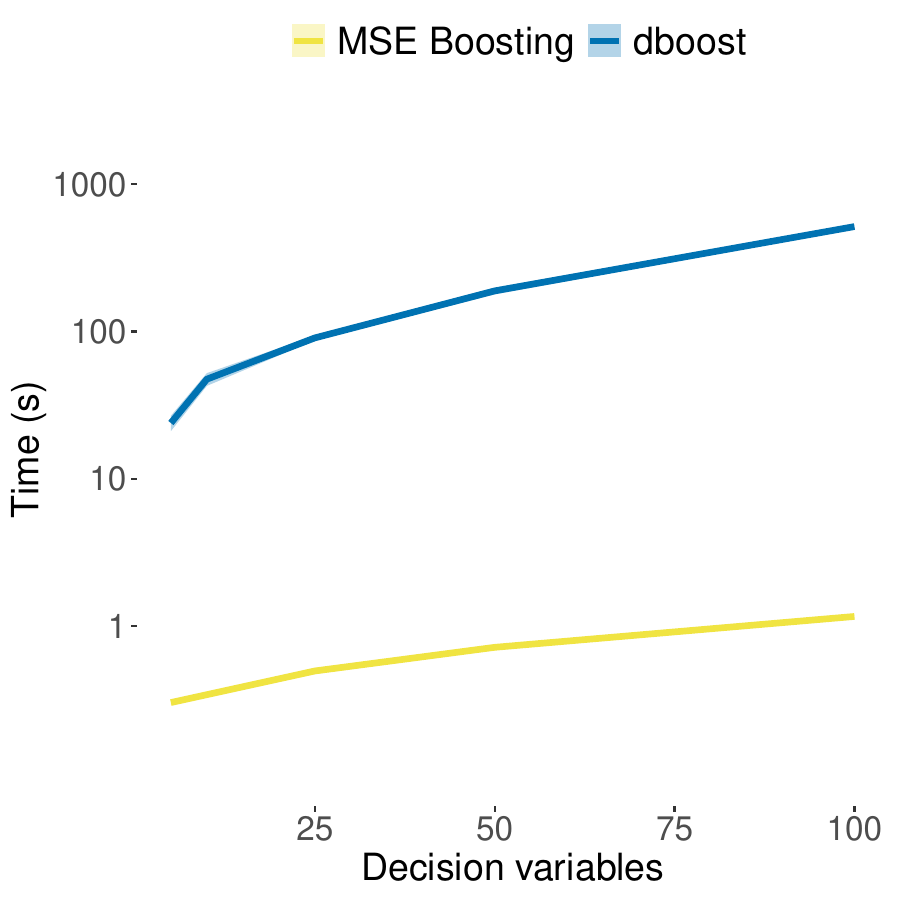}
    \caption{$m = 250$.}
  \end{subfigure}
    \begin{subfigure}[b]{0.32\linewidth}
    \includegraphics[width=\linewidth ]{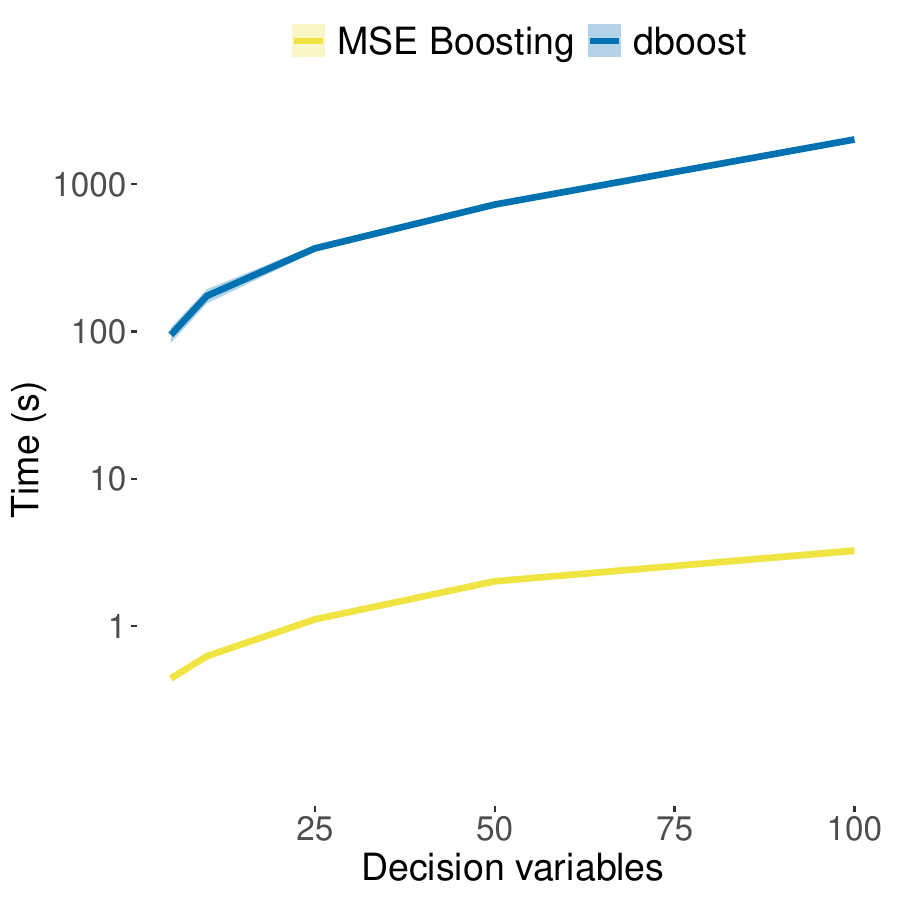}
    \caption{$m = 1000$.}
  \end{subfigure}
  \caption{Computation time and $95$\%-ile confidence interval of training \dboost~relative to MSE Boosting for the noisy portfolio optimization problem, evaluated over $10$ independent trials.  }
  \label{fig:comp_time}
\end{figure}

Figure \ref{fig:comp_time} reports the computation time (in seconds) and $95$\%-ile confidence interval of training  \dboost~relative to MSE Boosting for the noisy portfolio optimization problem, evaluated over $10$ independent trials. Here we consider training dataset sizes of $m  \in \{50, 250, 1000\}$ and decision variables in the range of $5-100$. In all cases, we limit the number of trees in the ensemble to $10$. In general we observe that training the \dboost~algorithm requires $20$x$- 600$x  more computation time than a traditional MSE Boosting algorithm. Indeed, for $m=1000$, the average computation time for training a MSE Boosting model is anywhere from $0.44 - 3.26$ seconds, whereas the average computation time for training the \dboost~model ranges from $95 - 2000$ seconds.  As expected, the computation time increases linearly as a function of the size of the training dataset. For example, when $m = 50$ and $d_z = 100$ the average computation time is $100$ seconds, whereas when $m=1000$ and $d_z = 100$, the average computation time is approximately $2000$ seconds; a $20$x increase. Improving the computational efficiency of \dboost~to support larger scale optimization problems is therefore an important area of future research.

\end{document}